# Cybernetic Environment: A Historical Reflection on System, Design, and Machine Intelligence

Zihao Zhang

University of Virginia, Virginia/USA · zz3ub@virginia.edu

**Abstract:** Taking on a historical lens, this paper traces the development of cybernetics and systems thinking back to the 1950s, when a group of interdisciplinary scholars converged to create a new theoretical model based on machines and systems for understanding matters of meaning, information, consciousness, and life. By presenting a genealogy of research in the landscape architecture discipline, the paper argues that landscape architects have been an important part of the development of cybernetics by materializing systems based on cybernetic principles in the environment through ecologically based landscape design. Landscape discipline has developed a design framework that provides transformative insights into understanding machine intelligence. The paper calls for a new paradigm of environmental engagement to understand matters of design and machine intelligence.

## 1 Introduction

In the past few years, the environmental management discourse has seen a myriad of research and practice which relies on technologies such as sensing networks, robotics, artificial intelligence (AI) and machine learning (ML) to regulate, control, and manage environmental processes. Principles in cybernetics have been introduced to envisage the environment as a system so that different strategies based on sensing-predict-control feedback loop can be deployed to influence the environment. First, sensing networks are introduced to collect various environmental data. Then these datasets are fed into ML algorithms to train AI systems or models to predict future scenarios and device control strategies that can be executed by cyber physical infrastructures. Finally, data of the updated environment are collected again and the sensing-predict-control mechanism comes to a full circle. These efforts give rise to a new paradigm of environmental engagement – cybernetic environment. However, in many cases, systems thinking and cybernetic control are assumed a priori in both theory and practice. What insights can we gain if a historical lens is taken to examine the origin and development of cybernetics? At the same time, these efforts have informed landscape designers to engage the environment in a new dimension. Ideas such as responsive landscapes and "third intelligence" have been developed to conceptualize machine intelligence in landscape design. How can these ideas contribute to the paradigm of cybernetic environment at large?

In order to respond to these two questions, this research first uses Katherine Hayles's scheme of "three waves of cybernetics" to present a brief history of the cybernetics movement and summarize major development in each wave. Through historical reflection, this paper argues that even though not fully recognized, landscape discipline has always been engaged with the field of cybernetics by materializing cybernetic systems in the physical environment. Constructing landscapes through ecology and systems thinking, the landscape architecture discipline has developed a systematic design framework to conceive machine intelligence in constructing the environment. This line of research needs to be distinguished from other versions of cybernetic imagination such as "smart city" and "smart environment" that envisage cybernetic technologies as a new type of digital infrastructure for optimizing environmental and urban processes. Rather, this paper traces a genealogy of research and practices which holds that machine intelligence is deeply ingrained in the environment and should be treated as a part of the reality. This line of thinking merits further research and can become a transformative force in the cybernetics environment discourse at large.

## 2　Rise of Cybernetic Environment: A New Paradigm

RangerBot is an underwater robot designed to protect the Great Barrier Reef from the overpopulated coral-eating crown-of-thorns (COT) starfish. This underwater vehicle relies on Random Forest Classifier, a machine learning technique for data classification, trained from underwater footage to track and detect COT starfish and kill them with lethal injection (DAYOUB et al. 2015). DroneSeed, a Seattle based company, uses drone swarms to plant and manage forests after disturbances such as wildfire. This drone-based reforestation strategy is claimed to be very efficient and six times faster than human planters. The flying path can be pre-programmed and optimized, and the drones can plant and manage trees semi-autonomously. Using these drone swarms, the company could propagate 80 acres of trees within 8 hours and reforest a post-wildfire site within 30 days.

In the realm of artificial intelligence (AI) and machine learning (ML), tech giants such as Facebook and Google have made promising results. Google has been using AI to predict all kinds of environmental-related events such as flooding and wind energy production. These AI systems are built with ML techniques that learn from data collected by sensing networks embedded in various environments. More inspiring results come from the research around deep reinforcement learning (DRL) techniques. DeepMind, which built AlphaGo that beat the best human Go player in 2016, has recently developed another AI system called AlphaStar that has reached grandmaster level in the real-time strategy video game StarCraft (SILVER et al. 2017, VINYALS et al. 2019). Many treat this experiment as a breakthrough in AI research because real-time strategy games such as StarCraft are infamously known for their "combinatorial action space, a planning horizon that extends over thousands of real-time decisions, and imperfect information" (VINYALS et al. 2019). After watching or playing with AlphaStar, many professional players reported that this gaming AI system has devised many new strategies that they can learn from, and they think AlphaStar has provided new ways to understand the game itself ("DeepMind's AlphaStar: A Grandmaster Level StarCraft 2 AI – YouTube" 2019). Even though a good number of exciting AI research revolves around video games, these AI systems are built with generality in mind. ML techniques and architectures used in these AI systems are general-purpose modules that are not specific to video games; rather they are general functions related to how *any* agent makes decisions in *any* given environment (VINYALS et al. 2019). This allows AI researchers to incorporate and recombine these general-purpose modules into other AI systems to tackle more complex phenomena that are beyond human comprehension, such as climate change.

However, there are at least two unquestioned and deeply related premises in most current cybernetic environment research. First, "system" is assumed a priori. Second, the underpinning framework revolves around sensing-predict-control protocol and research is trapped in the illusion of control through model-making. In other words, only because the environment is envisaged as a system we can develop feedback strategies to manage and control it. However, if we trace the development of cybernetics, it becomes clear that "system" itself is a mental model – a way of thinking or a paradigm – that humans construct to make sense of the world, bearing inevitable limitations and blind spots that prevent us from developing deeper understand of the cybernetic environment.

## 3  Three Waves of Cybernetics

The term cybernetics itself is largely misinterpreted because it is easy to associate "cyber" with digital and internet and overlook its original meaning. Cybernetics is derived from the Greek "kybernetes", meaning the person who steers a ship. WIENER (2000) in 1948 revived this concept in his book "*Cybernetics, or Control and Communication in the Animal and the Machine*" and developed cybernetics into an overarching approach to study organizational and control relations across systems. "To steer" is at the center of cybernetics since it studies how different systems use information, construct models, and exercise control actions to steer themselves towards their goals, or how we can instrumentalize "self-control" and "feedback loop" to nudge complex systems towards our desired direction. Cybernetics has given birth to many disciplines including computer science, robotics, and artificial intelligent (HEYLIGHEN & JOSLYN 2003). In over 70 years of development, its key concepts and modes of thinking have prevailed in academics and the general public at large. However, cybernetics has not become an independent discipline; instead, cybernetics has been mobilized among different fields over time. HAYLES in her seminal book "*How We Became Posthuman*" (1999) schematized the cybernetics movement into three waves and the frontier of cybernetics mobilized over different fields of research.

The first wave of cybernetics speaks to a series of interdisciplinary meetings from 1944 to 1953 known as the Macy Conferences on Cybernetics, which brought together many important post-war intellectuals including mathematicians Norbert Wiener, Claude Shannon, and John von Neumann, anthropologists Gregory Bateson and Margaret Mead, neurophysiologist Warren McCulloch, physicist and philosopher Heinz von Foerster, and psychiatrist W. Ross Ashby. Key concepts explored in the first wave were *homeostasis* and *self-regulating* systems. Homeostasis can be expressed by $f(x) = x$, in which the output of the function feeds back to the function itself as input. One classic example of a homeostatic system is a thermostat that controls room temperature. Cybernetic machines such as automatons and robots were built to test this theoretical framework. These machines are designed more than tools to fulfil some tasks, but more of physical experiments that materialize the imagined system. Because the first wave of cybernetics found direct applications in control systems, many practice-based scholars were quickly drawn away from the more theoretical based endeavors and formed more focused disciplines such as computer science, robotics, and artificial intelligence (HEYLIGHEN & JOSLYN 2003).

The second wave focused on second-order cybernetics, which is a recursive application of cybernetics on itself. During the Macy Conferences era, scholars studied how self-regulating systems construct internal models to steer their actions. However, scholars quickly realized that they, as observers, were also constructing models of the cybernetic systems: $f(x) = x$ is a mental model that observers construct to describe self-regulating systems. They felt the urge for a different framework that accounts for observers and their role in modeling systems. This issue was resolved by transfiguring observers into "observing systems" and treating them as part of the system that he/she studies (fig.1). Following this line of research, one of the key concepts in this wave was *autopoiesis* (self-production) developed by Chilean biologists

MATURANA & VARELA (1980). An autopoietic system, such as a human, uses input to reproduce its organization or relationships of its components. In a seminal paper, "What the Frog's Eye Tells the Frog's Brain" (LETTVIN et al. 1959), biologists and neuroscientists demonstrated that the eye of the frog does not capture an image and transmit a perfect copy to the brain for interpretation. Rather, four groups of nerve fibers operate on the image first before sending this highly organized information to the brain. Each group of fibers is responsible for one type of operation on the visual data and expresses the images in terms of movement of objects rather than the level of illumination. The consequence of such a finding flips how people usually understand the eye as a sensory organ that captures the image of the reality and sends the whole package of data to the brain to interpret. Instead, the brain reconstructs a "reality" based on what the eye selectively tells it. This reconstruction of reality is essential for the survival of the frog; the frog does not feed on flies but fly-sized moving objects that the eye helps the brain to see. Autopoiesis results in a highly constructivist epistemology: autopoietic systems, including humans, do not have direct interaction with outside reality but only access to the reality constructed by its own system operation. In a way, "system" is a concept that the human brain constructs to explain causal relationships in the environment around us to ensure our own survival.

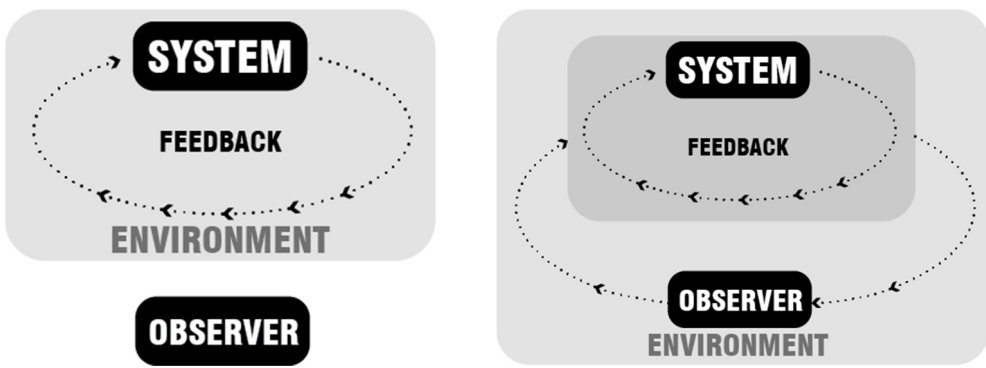

**Fig. 1:** First-order Cybernetics (Left) and Second-order Cybernetics (Right), illustration by Zihao Zhang

The third wave of research is an extension and development of second-order cybernetics in terms of life, intelligence, and consciousness. If cybernetics itself is a model that humans construct to understand any system, then system behavior becomes less linear but more unpredictable. HAYLES attributed the third wave of cybernetics in the research of artificial life in Santa Fe institute during the 1990s. The key concept investigated was *emergence*. Complexity can be achieved by applying simple rules through multi-agent simulation. The researchers developed a computer program that could replicate itself based on simple rules. The rules also allowed chances for mutation between generations. After running the simulation millions of times, the computer programs became an ecosystem of species. Every time the simulation started over, a completely different ecosystem of programs would emerge: sometimes with one dominant species and sometimes multiple species reaching a balance (NMCS4ALL 2013). The research provided a different insight into evolution and ecology, and challenged the linear ecological model that assumes a climax or stable condition of natural ecosystems.

Current cybernetic environment discourse largely revolves around the first-order cybernetics framework; the goal is to conceive the environment as a homeostatic system so that technologies can be developed to influence the environment through feedback loops. In other words, current research can be generalized into advances in three ends: 1) developing better sensing networks and protocols to collect more data and gain better understanding of the environment; 2) developing better ML algorithm and architecture to train models and more accurately predict the environment; and 3) developing better cyberphysical systems to devise control policies and control actions. However, these efforts lack a critical understanding that the environment as a system is merely a construction to explain control and human agency. Without critical concepts from second-order cybernetics and putting humans into the system, machine intelligence can only be envisaged as a means to extend human control rather than part of the reality that needs to be consciously designed.

## 4 Landscape Design and Cybernetics

Development of cybernetics and landscape theory and practice in the 20th century are not two parallel genealogies. Instead, we can identify many concepts in cybernetics that manifest in landscape design. Scholars have shown interest in building connections between cybernetics and landscape design. It is argued that Ian McHarg's design framework is in the first-order cybernetics that views the ecosystem as a homeostatic system, and the work of landscape architects is to fight for *entropy* that puts threats to this stable condition; whereas Lawrance Halprin's scoring system is an exploration of second-order cybernetics (the third wave) that focuses on uncertainty, chance, and emergence (LYSTRA 2014). Moreover, Toronto's Downsville urban park competition in the late 1990s marked a full manifestation of cybernetics and systems theory in landscape design discourse (CZERNIAK 2001). Almost all entries envisaged the urban park as a system in and of itself, and developed systemic strategies to direct the development of the landscape system with different scales of undecided factors to account for uncertainty over time (BERRIZBEITIA 2001). Time becomes an important factor and landscape is considered process rather than result. For example, the entry by James Corner, Stan Allen & Nina Mari Lister adopted an that complexity can be created through ecological processes (REED & LISTER 2014a, 2014b). CHARLES WALDHEIM (2006) used "strategies of indeterminacy" to conceptualize adaptive management framework, and designed a landscape system that can evolve so landscape practices in the 1990s to 2000s. In the past three decades, landscape discipline has developed a design framework that focuses on emergence, indeterminacy, and uncertainty.

Deeply rooted in this landscape design framework, landscape architects start to conceptualize machine intelligence and explore design in a new dimension. In this vein of research, CANTRELL & HOLTZMAN (2016) foregrounded responsive landscapes framework and articulated sensing-processing-actuating feedback loop to approach design. Responsive technologies are not conceived as a layer on top of but as a network that is deeply embedded in the environment. Laden with intelligence, landscapes can evolve over time, respond to various inputs, and adapt to different scenarios. ZHANG & BOWES (2019) have provided empirical evidence which shows that machine intelligence has been deeply involved in environmental processes and become an important player in the decision-making processes of environmental management. Posthumanism is introduced to conceptualize machine intelligence; designers should overcome anthropocentrism when theorizing and applying ML and AI in the land-scape discipline (ZHANG & BOWES 2019). In a similar vein, devising the concept of "third intelligence", CANTRELL & ZHANG (2018) theorized that machines are one of many intelligent agents, including humans, that co-evolve and co-produce the shared environment. Rather than another type of control mechanism that is introduced to the environment through which humans exercise more agency, "third intelligence" holds that machine intelligence can be deeply ingrained in the environment. In order to design with different types of agent/intelligence, ROBINSON & DAVIS (2018) have argued for new kinds of interface between designers and landscape.

Practices remain speculative but merit further exploration. In a thought experiment, scholars imagined a DRL based machine "wildness creator", just like AlphaGo and AlphaStar, that can devise environmental management strategies and create places that are beyond human comprehension (CANTRELL et al. 2017). Rather than envisioning DRL based AI system as "superhumans" who can predict the environment in human discourse, "wildness creator" is conceptualized as a different type of intelligence that understands the environment differently than humans.

In a research design studio, students designed and built a sensing station on site. The sensing station is conceived as an interface of the site system that provides continuous contextual information to designers who then develop strategies over time (OSBORN 2018). Rather than a sensing network that tries to provide "objective" data for developing control strategies, the sensing station evolves into part of the reality to which the designer must respond.

Moreover, designer LEIF ESTRADA (2018) proposed sensing networks, data processing and actuating systems in a fluvial environment, and through rigorous experiments on a geomorphology table, the coupled technical-natural system shows a level of real-time response that is beyond human capability.

In a similar framework, in order to interact with a remote cryosphere environment in microscale, designer GONZALEZ RAMIREZ (2016) has designed a system with a central mind and distributed actuators (bodies) deeply embedded in the cryosphere environment. The mind can cast a vast array of actuating policies across the bodies and evaluate the policies based on discrepancies between projected scenarios and actual responses from the environment. Based on a series of evaluations, the system can adjust policies over time and evolve with the cryosphere environment.

## 5    Conclusion: A Fourth Wave of Cybernetics

Cybernetics is not at all about technologies but about the study of communication and control across different entities including humans, machines, and other nonhuman entities. In other words, technology and machine in the cybernetics framework do not serve as a means to an end or tools that help to optimize processes. Instead, a machine is only a model for understanding how a system functions, and a lens for examining how different entities interact. The application of technologies and scientific research are not the core interests in the field of cybernetics, rather they serve as empirical evidence for scholars to ask deeper questions about information, life, evolution, and consciousness. Consequently, when considering emerging technologies such as AI and ML in the context of environmental design and management, we need to bypass the urge to find useful applications in these tools. Rather, a very important lesson from studying the development of cybernetics is that the cybernetic envi-ronment paradigm can provide us with another theoretical framework for examining the environment in a different light. In a way, the environment that designers have to engage today is intrinsically different from the environment that designers had engaged decades ago. The environment is laden with machine intelligence that has been introduced by humans but not fully comprehended by us. We should recognize that machine intelligence is not a thin layer that is added to the environment to extend the regime of human control. Rather, machine intelligence is deeply engrained in the environment and becomes part of the reality that we have to understand and design with. We have to develop and embrace a new theoretical framework to understand matters of machine intelligence and landscape design.

Landscape architecture has always been part of cybernetics by materializing systems and building physical manifestations of cybernetic principles. Deeply rooted in ecology and systems thinking, the landscape discipline has developed a design framework that transforms how intelligence is conceived and how the machine is envisaged in constructing the environment. Thus, this paper calls for a fourth wave of cybernetics that is rooted in the study of the cybernetic environment. In addition, and more importantly, landscape architects have the knowledge and sensibility to consider and design how different entities co-evolve and co-produce the shared environment.